\documentclass{article}
\usepackage{ijcai22}

\usepackage{times}
\usepackage{soul}
\usepackage{url}
\usepackage[hidelinks]{hyperref}
\usepackage[utf8]{inputenc}
\usepackage[small]{caption}
\usepackage{subcaption,graphicx}
\usepackage{amsmath, amsfonts}
\usepackage{amsthm}
\usepackage{booktabs}
\usepackage{algorithm}
\usepackage{algorithmic}
\usepackage{xcolor}
\usepackage{textcomp}
\usepackage{float}
\usepackage{multicol}
\usepackage{mathtools}
\usepackage[export]{adjustbox}
\usepackage{mathalfa}
\usepackage{color, colortbl}
\usepackage{hyperref}
\usepackage{breqn}
\usepackage{csquotes}
\hypersetup{colorlinks,linkcolor={blue},citecolor={blue}}  
\definecolor{LightCyan}{rgb}{0.88,1,1}

\newcommand{\etal}{{\textit{et al.}}}

\title{Beckman Defense}

\author{
    A V Subramanyam
    \affiliations
    IIITD
    \emails
    subramanyam@iiitd.ac.in
}

\DeclareMathOperator*{\argmax}{arg\,max}
\DeclareMathOperator*{\argmin}{arg\,min}

\begin{document}

\maketitle
\begin{abstract}
Optimal transport (OT) based distributional robust optimisation (DRO) has received some traction
in the recent past. However, it is at a nascent stage but has a sound potential in robustifying the deep learning models.  Interestingly, OT barycenters demonstrate a good robustness against adversarial attacks. Owing to the computationally expensive nature of OT barycenters, they have not been investigated under DRO framework. In this work, we propose a new barycenter, namely Beckman barycenter, which can be computed efficiently and used for training the network to defend against adversarial attacks in conjunction with adversarial training. We propose a novel formulation of Beckman barycenter and analytically obtain the barycenter using the marginals of the input image. We show that the Beckman barycenter can be used to train adversarially trained networks to improve the robustness. Our training is extremely efficient as it requires only a single epoch of training. Elaborate experiments on CIFAR-10, CIFAR-100 and Tiny ImageNet demonstrate that training an adversarially robust network with Beckman barycenter can significantly increase the performance. Under auto attack, we get a a maximum boost of 10\% in CIFAR-10, 8.34\% in CIFAR-100 and 11.51\% in Tiny ImageNet. Our code is available at https://github.com/Visual-Conception-Group/test-barycentric-defense.
\end{abstract}

\section{Introduction}
\label{sec:intro}
Optimal mass transport (OT), originally proposed by Monge in his seminal work \cite{monge1781memoir}, has gathered a widespread interest in the field of learning representations. The original deterministic OT problem was later relaxed by Kantorovich \cite{kantorovich1942transfer} and considered a probabilistic transport problem. This formulation seeks solution for the optimal transport plan which can transport mass between two measures by incurring the minimum cost and is solved using a linear program. The modern day OT is also attributed to the phenomenal work of Kantorovich. Following the OT theory, barycenters in Wasserstein space was proposed by Agueh and Carlier in their remarkable work \cite{agueh2011barycenters}. Further, using entropic regularization \cite{cuturi2013sinkhorn}, a fast method of computing barycenters was proposed by Cuturi and Doucet \cite{cuturi2014fast}. Recent works addresses the challenge of computational complexity of barycenters using neural networks \cite{lacombe2021learning}. In this work, we investigate the barycenters towards robust learning of deep learning models.

{D}{eep} learning systems have shown impressive performance in various applications. However, these systems are vulnerable to adversarial perturbations \cite{wong2020fast}, \cite{croce2020reliable}, \cite{xie2019improving}. In order to counter these attacks, several defense mechanisms have also been proposed. In one of the early works, Szegedy \etal~\cite{szegedy2013intriguing} formulated the adversarial attack as an optimization problem and obtained the adversarial sample using L-BFGS. Several adversarial attacks have been proposed since Szegedy' work \cite{goodfellow2014explaining,kurakin2016adversarial}. On the other hand, strong defense measures have been studied in \cite{madry2017towards}, \cite{theagarajan2019shieldnets}, \cite{wong2020fast}, \cite{rebuffi2021fixing}. 
 

 \begin{figure}[h]
     \centering
    \includegraphics[width=.45\textwidth]{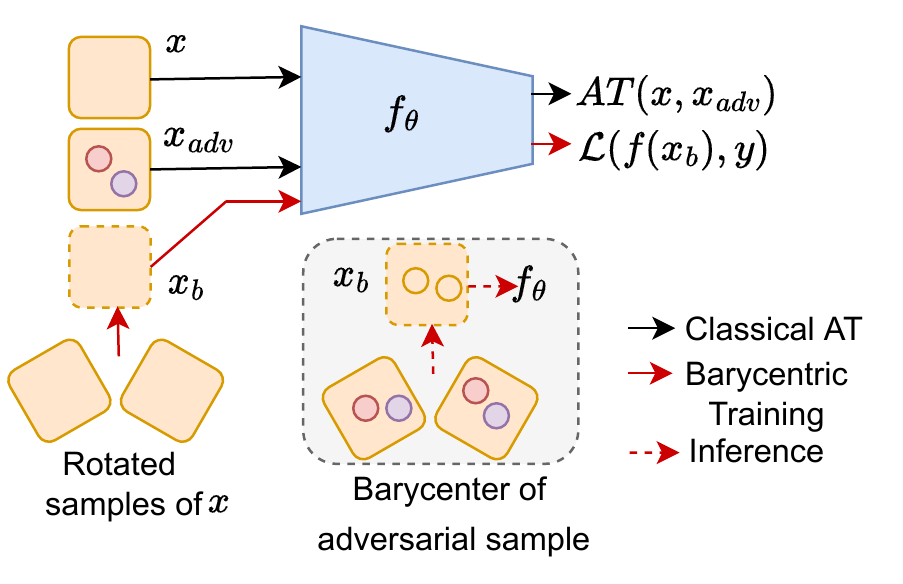}
     \caption{Illustration: Classical defense methods use Adversarial Training (AT) as a major defense technique. Our method obtains barycenter from rotated inputs and uses them for training the model using a cross-entropy loss. During inference time also we compute barycenter of the given sample. The dashed boundary of barycenter indicates that the barycenter is close to input samples in terms of appearance but there are some differences. In the computation of barycenter of adversarial sample, the barycenter shows the changes in same color as that of the background to imply that barycenter suppresses the adversarial noise.}
     \label{fig:teaser}
\end{figure}

In the field of adversarial attacks and defense, $l_p$ space has been extensively studied. However, only a few works investigate attacks under OT framework \cite{wong2019wasserstein}, \cite{li2021internal}. There are even fewer works which investigate robustness using OT theory \cite{kwon2020principled}, \cite{subramanyam2022barycentric}. Distinct from these works, we first introduce Beckman barycenter, a concept analogous to Wasserstein barycenter. We use proximal operator methods to solve for the barycenter. The barycenters obtained from the clean samples are used to train a pretrained adversarially robust network. We note that in the absence of adversarial samples in the training, the model would give a better clean accuracy but will suffer in terms of adversarial accuracy. Therefore, we use a pre-trained adversarially robust network to overcome this challenge. An abstract illustration of our method is given in Figure \ref{fig:teaser}. 

Beckman barycenter is obtained from input marginals via a non-linear interpolation. The input marginals are linearly transformed versions of the input and thus interfere with the adversarial noise. Using these marginals the barycenter generates a sample which is similar in appearance to the input and is  closer in terms of class label. Thus, the class label is preserved when the input is a clean sample, whereas, the adversarial noise gets suppressed when the input is an adversarial sample. Further, the network needs to be trained with barycenter of clean samples so as to correctly classify them. However, this training is cheap as a single epoch is sufficient. We prove our hypothesis using extensive qualitative and quantitative experiments.

\section{Related Works}
\textbf{Adversarial Attacks} Given an adversarial sample $\mathbf{x}$ with label $y$, a target network $f$ parameterized by $\theta$, the adversary tries to find $\mathbf{x}_{adv}$ by adding an adversarial noise such that the prediction $f_{\theta}(\mathbf{x}_{adv}) \neq f_{\theta}(\mathbf{x}) = y$. Some of the robust attacks are iterative FGSM \cite{kurakin2016adversarial}, PGD \cite{madry2017towards}, Carlini
and Wagner attacks \cite{carlini2017towards}, Jacobian based attack \cite{papernot2016limitations}, physical attack Athalye \cite{athalye2018synthesizing}, and Autoattack \cite{croce2020reliable}. These attacks are primarily focused in $l_p$ domain. 

\noindent \textbf{Adversarial Defense} 
In response to adversarial attacks, several defenses have been proposed. One of the best defense approach is adversarial training \cite{szegedy2013intriguing}, \cite{goodfellow2014explaining}, \cite{moosavi2016deepfool}. Madry \etal~\cite{madry2017towards} formally studied adversarial training and proposed that such training allows network to defend well against first order adversary. Adversarial logit pairing uses a pair of logits from clean and adversarial examples to defend against adversarial samples \cite{kannan2018adversarial}. TRADES \cite{zhang2019theoretically} prove the bounds based on regularization term which minimizes the difference in prediction between clean and adversarial examples. In \cite{wong2020fast}, authors proposed to effectively combine FGSM and random initialization to demonstrate better adversarial training. RST \cite{carmon2019unlabeled} propose a self-training technique using unlabelled samples to improve the robustness. Observing the correlation between flatness of weight loss landscape and adversarial robustness, Wu \etal~proposed adversarial weight perturbation (AWP) to regularize the flatness of weight loss \cite{wu2020adversarial}. On similar lines, \cite{yu2022robust} propose a criterion called Loss Stationary Condition (LSC) for constrained perturbation, which regulates the weight perturbation to prevent overfitting. LBGAT \cite{cui2021learnable} constrains the logits of a robust model, trained with adversarial examples, to be similar to the logits of a clean model trained on natural data. 

While adversarial training uses all the samples, many techniques propose that naively using adversarial samples in adversarial training is not efficient. This primarily involves training the model with a weak attack first, and then gradually increasing the strength of the adversary - CAT \cite{cai2018curriculum}, DART \cite{wang2019dynamic}, MART \cite{wang2019improving}, FAT \cite{zhang2020attacks}. Aforementioned methods rely on pre-determined attack parameters for adversarial sample generation. However, this restricts the model's robustness. To address this issue, LAS-AT \cite{jia2022adversarial} propose a framework for adversarial training that introduces the notion of learnable attack strategy. It is composed of two components: a target network that uses adversarial examples for training to improve robustness, and a strategy network that produces attack strategies to control adversarial sample generation. In similar spirit, A2 \cite{xu2022a2} and \cite{cheng2022cat} have also been proposed. A classical review of defense methods can be obtained in \cite{ijcai2021p591}.

In a parallel line of defense works, input purification has also been explored. At the test time, these techniques try to remove the adversarial noise \cite{shi2021online}, TRADES$_{\text{SSL}}$ \cite{mao2021adversarial}, Hedge$_{\text{RST}}$ \cite{wu2021attacking}. Score based generative models such as \cite{yoon2021adversarial} and \cite{nie2022diffusion} have also been used to purify the images before sending them for classification.


Our work is inspired from two different theories, namely, OT barycenters and distributional robust optimization. We discuss these theories in the following.

\noindent \textbf{Wasserstein Barycenter} In the following we discuss Wasserstein distance and barycenter. Given probability distributions, $ \boldsymbol{\mu}_1, \boldsymbol{\mu}_2$ $\in$ $\Omega$, the Wasserstein distance is defined as, 
\begin{align}
\mathcal{W}(\boldsymbol{\mu}_1, \boldsymbol{\mu}_2) = \inf_{\Omega \times \Omega} c(x,y) \pi(x,y) dx dy, \label{eq:wbary} \\
s.t. \;\int_{\Omega} \pi(x,y) dx = \boldsymbol{\mu}_1(x), \int_{\Omega} \pi(x,y) dy = \boldsymbol{\mu}_2(y), \nonumber
\end{align}
where the cost matrix $c(x,y) = \lVert x - y\rVert_1$ and $\pi$ denotes the transport plan. This is also known as Earth Mover' Distance (EMD). This form is also used to compute barycenter \cite{cuturi2016smoothed} wherein the summation of Wasserstein distance between the barycenter and each input marginal is considered. However, barycenters are costly to compute and the best known complexity scales exponentially with the number of marginals \cite{fan2022complexity}. 

EMD can also be represented as dual of the dual of Eq \ref{eq:wbary} in variational form popularly introduced by Beckman \cite{beckmann1952continuous}, \cite{li2018parallel}, \cite{lee2020unbalanced},
\begin{align}
\mathcal{W}(\boldsymbol{\mu}_1, \boldsymbol{\mu}_2) =\inf_{\mathbf{M}} \int_{\Omega} \lVert \mathbf{M} \rVert  \label{eq:EMD}\\
  s.t.\; \text{div}(\mathbf{M}) + \boldsymbol{\mu}_1 - \boldsymbol{\mu}_2 = 0 \nonumber \\ \nonumber
  \mathbf{M}.\mathbf{n} = 0\; \forall x \in \partial \boldsymbol{\Omega}; n \; \text{is normal to}\; \partial  \boldsymbol{\Omega} \nonumber
\end{align}
Under appropriate discretization, $\mathbf{M} = (\mathbf{M}_x, \mathbf{M}_y)$, $\mathbf{M} \in \mathcal{R}^{n \times 2}$ is flux vector satisfying zero flux boundary conditions. $\boldsymbol{\mu}_1, \boldsymbol{\mu}_2 \in \mathcal{R}^{n}$, and,
\begin{equation}
\text{div}(\mathbf{M}) = (\mathbf{M}_x[i,j] - \mathbf{M}_x[i-1,j]) + (\mathbf{M}_y[i,j] - \mathbf{M}_y[i,j-1]) \nonumber
\end{equation}
and the zero-flux boundary conditions mean that $\mathbf{M}_x[i,j] = \mathbf{M}_y[i,j] = 0$ outside the boundary. Eq \ref{eq:EMD} is favorable compared to Eq \ref{eq:wbary} as it reduces the complexity from $\mathcal{O}(n^2)$ to $\mathcal{O}(n)$ \cite{li2018parallel}. Motivated by the recent developments of OT barycenters, we make use of Eq \ref{eq:EMD} to propose Beckman barycenter as they can be efficiently solved using well known techniques like \cite{goldstein2009split}, \cite{chambolle2011first}.


\noindent \textbf{DRO} One of the influential works in DRO was proposed by Scarf \cite{scarf1957min}. Following this work, significant research has been done in this field \cite{ben2009robust}, \cite{duchi2021statistics}, \cite{staib2017distributionally}. DRO aims to address the problem of uncertainty or shift in the data distribution that can arise due to measurement errors and admits a solution for the worst case scenario. Let $\mathcal{L}(\boldsymbol{\theta}, \mathbf{x})$ be the loss function where $\boldsymbol{\theta}$ are network parameters. Then, DRO solves for,
\begin{equation}
\inf_{\boldsymbol{\theta}} \sup_{Q \in \mathcal{Q}} \mathbb{E}_{\mathcal{Q}} \mathcal{L}(\boldsymbol{\theta}, \mathbf{x})
\label{eq:DRO}
\end{equation}
Here, $\mathcal{Q}$ is the distribution against which DRO minimizes the loss. For instance, $\mathcal{Q}$ can be considered as a distribution set which contains perturbations of input samples $\mathbf{x}$. Here we note that adversarial training can be considered to be a specific instance of DRO wherein the distribution $\mathcal{Q}$ is drawn from adversarial samples. In our case, we consider the barycenters as the samples drawn from the distribution $\mathcal{Q}$ and thus provide robustness against perturbed samples.

\subsection{Proposed Algorithm}
In this work, we propose a novel Beckman barycenter formulation and derive the barycenter analytically. We use the barycenter to demonstrate that it can be applied for adversarial defense. We first obtain the barycenter using the marginals from the given input image and then train the network using barycenter. 

While OT barycenters are a good choice for the distribution $\mathcal{Q}$ in Eq \ref{eq:DRO}, computing OT barycenter suffers from high complexity and exponentially increases with the number marginals \cite{fan2022complexity}. To counter this high complexity challenge, we first discuss an analogous barycenter problem by building upon the formulation given in Eq \ref{eq:EMD},
\begin{align}
\inf_{\substack{\mathbf{M}_1,\mathbf{M}_2\\ \mathbf{r}_1,\mathbf{r}_2,\boldsymbol{\mu}}} \lVert \mathbf{M}_1 \rVert_{2,1} + \lVert \mathbf{M}_2 \rVert_{2,1} + \alpha(\lVert \mathbf{r}_1 \rVert_1 \label{eq:EMD-barycenter1} \\ + \lVert \mathbf{r}_2 \rVert_1) + \beta\lVert \boldsymbol{\mu} \rVert_1 \nonumber \\ \nonumber
  s.t.\; \text{div}(\mathbf{M}_1) + \boldsymbol{\mu}_1 - \boldsymbol{\mu} = \mathbf{r}_1 \nonumber\\ \nonumber
  \text{div}(\mathbf{M}_2) + \boldsymbol{\mu}_2 - \boldsymbol{\mu} = \mathbf{r}_2 \nonumber\\ \nonumber
\end{align}
where, $\mathbf{r}_1, \mathbf{r}_2, \boldsymbol{\mu} \in \mathcal{R}^n$.
Our formulation is loosely inspired from the Beckman OT formulation that is given in \cite{li2018parallel}, \cite{lee2020unbalanced}.
There are notable changes in Eq \ref{eq:EMD-barycenter1} from Eq \ref{eq:EMD}. First we solve for Beckman barycenter $\boldsymbol{\mu}$ in addition to other variables. Similar to Wasserstein barycenter which acts as a representative of marginals using Wasserstein metric, the Beckman barycenter $\boldsymbol{\mu}$ minimizes the flux with respect to input marginals $\boldsymbol{\mu}_1$ and $\boldsymbol{\mu}_2$. In our experiments, these marginals are obtained by rotating the input image with $\pm4^{\circ}$. Second, the variables $\mathbf{r}_1$ and $\mathbf{r}_2$ allow the mass to be created or destroyed \cite{lee2020unbalanced} and the regularization over $\mathbf{r}_1$, $\mathbf{r}_2$ and $\boldsymbol{\mu}$ ensure that these variable do not take arbitrarily large values. Third, Eq \ref{eq:EMD-barycenter1} can be easily converted to Lagrange formulation and solved in linear time using primal-dual method of Chambolle and Pock \cite{chambolle2011first}. 

In order to make the objective strongly convex, we first apply proximal operators. The $l_2$ regularizer makes the objective strongly convex.
Using the proximal operator,
\begin{align}
\inf_{\substack{\mathbf{M}_1,\mathbf{M}_2,\mathbf{r}_1\\ \mathbf{r}_2, \boldsymbol{\mu}_1',\boldsymbol{\mu}_2',\boldsymbol{\mu}}}  \lVert \mathbf{M}_1 \rVert_{2,1} + \lVert \mathbf{M}_2 \rVert_{2,1} + \alpha(\lVert \boldsymbol{r}_1 \rVert_1 \label{eq:EMD-barycenter2}\\ + \lVert \boldsymbol{r}_2 \rVert_1) + \frac{1}{2\rho}(\lVert \boldsymbol{\mu}_1' - \boldsymbol{\mu}_1 \rVert_2 + \lVert \boldsymbol{\mu}_2' - \boldsymbol{\mu}_2 \rVert_2) +\beta\lVert \boldsymbol{\mu} \rVert_1\nonumber\\
  s.t.\; \text{div}(\mathbf{M}_1) + \boldsymbol{\mu}_1' - \boldsymbol{\mu} = \boldsymbol{r}_1 \nonumber \\ \nonumber
  \text{div}(\mathbf{M}_2) + \boldsymbol{\mu}_2' - \boldsymbol{\mu} = \boldsymbol{r}_2 \nonumber
\end{align}

The Lagrangian of Eq \ref{eq:EMD-barycenter2} is given as,

\begin{align}
\inf_{\substack{\mathbf{M}_1,\mathbf{M}_2,\mathbf{r}_1\\ \mathbf{r}_2, \boldsymbol{\mu}_1',\boldsymbol{\mu}_2',\boldsymbol{\mu}}}  \lVert \mathbf{M}_1 \rVert_{2,1} + \lVert \mathbf{M}_2 \rVert_{2,1} + \alpha(\lVert \boldsymbol{r}_1 \rVert_1 \label{eq:EMD-barycenter3}\\ + \lVert \boldsymbol{r}_2 \rVert_1) + \frac{1}{2\rho}(\lVert \boldsymbol{\mu}_1' - \boldsymbol{\mu}_1 \rVert_2 + \lVert \boldsymbol{\mu}_2' - \boldsymbol{\mu}_2 \rVert_2)+\beta\lVert \boldsymbol{\mu} \rVert_1\nonumber \\\nonumber
+\sum_i \langle \boldsymbol{\lambda}_i, \text{div}(\mathbf{M}_i) + \boldsymbol{\mu}_i' - \boldsymbol{\mu} - \boldsymbol{r}_i \rangle \nonumber 
\end{align}

Eq \ref{eq:EMD-barycenter3} can be solved using first-order primal dual method of Chambolle and Pock \cite{chambolle2011first}\footnote{We use similar notations to that of \cite{li2018parallel}, \cite{chambolle2011first} for consistency and simplicity.}. 
\begin{gather*}
\mathbf{M}_i^{t+1} \leftarrow \argmin_{\mathbf{M}_i} \lVert \mathbf{M}_i \rVert_{2,1} + \langle \boldsymbol{\lambda}_i,  \text{div}(\mathbf{M}_i) + \boldsymbol{\mu}_i' - \\ \boldsymbol{\mu} - \mathbf{r}_i \rangle
+ \frac{1}{2\tau_1} \lVert \mathbf{M}_i - \mathbf{M}_i^t \rVert_2 \;\;\;\;\forall i = \{1, 2\} \nonumber\\
{\boldsymbol{\mu}^{\prime}_i}^{t+1}  \leftarrow \argmin_{{\boldsymbol{\mu}^{\prime}_i}}  \frac{1}{2\tau_1}(\lVert \boldsymbol{\mu}_i' - \boldsymbol{\mu}_i \rVert_2)  + \langle \boldsymbol{\lambda}_i, \boldsymbol{\mu}_i' \rangle \\ +\frac{1}{2\tau_1} \lVert \boldsymbol{\mu}_i' - {\boldsymbol{\mu}^{\prime}_i}^{t} \rVert_2\\ \nonumber
\mathbf{r}_i^{t+1} \leftarrow \argmin_{\mathbf{r}_i} \alpha \lVert \boldsymbol{r}_i \rVert_1 + \langle \boldsymbol{\lambda}_i^t, -\mathbf{r}_i  \rangle + \frac{1}{2\tau_1} \lVert \boldsymbol{r}_i - \boldsymbol{r}_i^t \rVert_2\\ \nonumber
\boldsymbol{\mu}^{t+1} \leftarrow \argmin_{\boldsymbol{\mu}} \lVert \boldsymbol{\mu} \rVert_1 + \langle \boldsymbol{\lambda}_i^t, -\boldsymbol{\mu}  \rangle + \frac{1}{2\tau_1} \lVert \boldsymbol{\mu} - \boldsymbol{\mu}^t \rVert_2\\ \nonumber
\boldsymbol{\lambda}_i^{t+1} \leftarrow \argmax_{\boldsymbol{\lambda}_i} \langle \boldsymbol{\lambda}_i, \boldsymbol{\kappa}^{t+1}   \rangle - \frac{1}{2\tau_2}\lVert \boldsymbol{\lambda}_i - \boldsymbol{\lambda}_i^t\rVert_2,
\end{gather*}
where, $\boldsymbol{\kappa}^{t+1} = 2(\text{div}(\mathbf{M}_i)^{t+1} +{\boldsymbol{\mu}^{\prime}_i}^{t+1} - \boldsymbol{r}_i^{t+1})-(\text{div}(\mathbf{M}_i)^t + {\boldsymbol{\mu}^{\prime}_i}^{t} - \boldsymbol{r}_i^t)$.

We now discuss the solution of each individual optimization. 

\textbf{Solving for} $\mathbf{M}_i$: The rows $\mathbf{m}_{ij}$ of $\mathbf{M}_i$ can be expressed and solved using $l_{21}$ norm shrinkage operator,
\begin{equation}
\mathbf{m}_{ij}^{t+1} \leftarrow \text{shrink}_{\tau_1}^{l_2}(\mathbf{m}_{ij}^t - \tau_1 \text{div}^*(\boldsymbol{\lambda}_i^t)_j)
\label{eq:solveforM}
\end{equation}
Here, $\text{div}^*$ denotes the adjoint of div operator, and $\text{shrink}_{\tau_1}^{l_2}{\boldsymbol{\eta}} = \max(\lVert \boldsymbol{\eta} \rVert_2 - \tau_1, 0) \odot \frac{\boldsymbol{\eta}}{(\lVert \boldsymbol{\eta} \rVert_2)}$.  \enquote{$\odot$} denotes the Hadamard product.
    
\textbf{Solving for} $\boldsymbol{\mu}_i'$:
\begin{equation}
\centering
{\boldsymbol{\mu}^{\prime}_i}^{t+1} \leftarrow \max\{0, \frac{\rho\tau_1}{1+ \rho\tau_1}\boldsymbol{\mu}_i' + \frac{1}{1+\rho \tau_1}({\boldsymbol{\mu}_i'}^t - \tau_1 \boldsymbol{\lambda}_i^t) \},
\end{equation}

\textbf{Solving for} $\mathbf{r}_i$:
We use an $l_1$ shrinkage operator.
\begin{equation}
\mathbf{r}_{i}^{t+1} \leftarrow \text{shrink}_{\alpha \tau_1}^{l_1}(\mathbf{r}_{i}^t +  \tau_1 \boldsymbol{\lambda}_i^t)
\end{equation}

Here, $\text{shrink}_{\alpha\tau_1}^{l_1}(\boldsymbol{\eta}) = \text{sign}(\boldsymbol{\eta}) \odot \max(\lVert \boldsymbol{\eta} \rVert - \alpha\tau_1, 0)$.

\textbf{Solving for barycenter} $\boldsymbol{\mu}$:
\begin{equation}
    \boldsymbol{\mu}^{t+1} \leftarrow \text{shrink}_{\beta \tau_1}^{l_1}(\boldsymbol{\mu}^t +  \tau_1 (\boldsymbol{\lambda}_1^t+\boldsymbol{\lambda}_2^t))
\end{equation}

\textbf{Solving for} $\boldsymbol{\lambda}$:
\begin{equation}
\boldsymbol{\lambda}_i^{t+1} \leftarrow \boldsymbol{\lambda}_i^{t} + \tau_2 \boldsymbol{\kappa}^{t+1}
\label{eq:solveforlambda}
\end{equation}

\subsection{Toy example}
\begin{figure}[h]
     \centering
     \begin{subfigure}{.12\textwidth}
        \includegraphics[width=\textwidth]{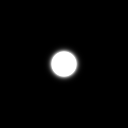}
     \end{subfigure}   
     \begin{subfigure}{.12\textwidth}
        {\includegraphics[width=\textwidth]{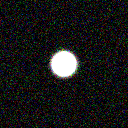}}
     \end{subfigure}   
     \begin{subfigure}{.12\textwidth}
        {\includegraphics[width=\textwidth]{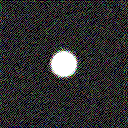}}
     \end{subfigure}     
    \begin{subfigure}{.12\textwidth}
         \includegraphics[width=\textwidth]{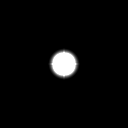}
         \caption{Clean}
    \end{subfigure}
    \begin{subfigure}{.12\textwidth}
         \includegraphics[width=\textwidth]{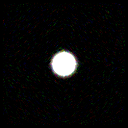}
        \caption{Random}
     \end{subfigure}   
     \begin{subfigure}{.12\textwidth}
         \includegraphics[width=\textwidth]{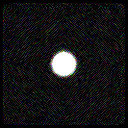}
         \caption{Adversarial}
     \end{subfigure}        
        \caption{Top: Clean image, noisy image, adversarial image. Bottom: Barycenter of clean image, noisy image, adversarial image.} 
        \label{fig:clean-and-adversarial}
\end{figure}

We demonstrate the barycenter computation using a Gaussian image in Figure \ref{fig:clean-and-adversarial}. The 
 barycenter of clean samples, sample with random noise and adversarial sample are shown. As we see, for the clean case the barycenter is very similar to that of the original image. In the second column where random  noise is added, the barycenter reduces the noise. Similar effect is also seen for the case where adversarial noise is present. This indicates that non-linear interpolation of barycenter suppresses the adversarial noise. 

\subsection{Training}
Let a model be given by $f_{\boldsymbol{\theta}}$, the barycenter of clean samples be denoted by $\mathbf{x}$ and its labels as $y$. We then optimize the following loss
\begin{equation*}
    \argmin_{\boldsymbol{\theta}}\frac{1}{n}\sum_{i=1}^n \mathcal{L}_{CE}(f_{\boldsymbol{\theta}}(\mathbf{x}_i), \mathbf{y}_i)
\end{equation*}
where $\mathcal{L}_{CE}$ is the cross-entropy loss. We would like to emphasize that we do not perform adversarial training. Instead we use an adversarially pretrained model. Thus, $f_{\boldsymbol{\theta}}$ is an adversarial robust model and our training further enhances the robustness. We also note that this optimization falls under DRO as the samples used are barycenters which belong to the distribution $\mathcal{Q}$.

\subsection{Theoretical analysis}
We first present a convergence analysis of Eq \ref{eq:EMD-barycenter3}.

\textbf{Theorem 1.}
Let $\tau_1 \tau_2(\lambda_{\max}(\boldsymbol{\nabla}^2)+3) < 1$, where $\lambda_{max}(\boldsymbol{\nabla}^2)$ denotes the largest eigenvalue of discrete Laplacian operator $\boldsymbol{\nabla}^2 = \mathbf{D}\mathbf{D}^{\top}$, where $\mathbf{D}$ is the matrix representing div operator. Then, the iterations 
$\mathbf{M}^t_i, {\boldsymbol{\mu}^{\prime}}^t_i, \boldsymbol{\mu}^t, \mathbf{r}^t_i, \boldsymbol{\lambda}^t_i$ converge to the saddle point solution of the Lagrangian $\mathbf{M}^*_i, \boldsymbol{\mu}^*_i, \boldsymbol{\mu}^*,\mathbf{r}^*_i, \boldsymbol{\lambda}^*_i$.

Proof: 
Let $\mathbf{u}=\{\mathbf{M}_1,\mathbf{M}_2, \boldsymbol{\mu}_2, \boldsymbol{\mu}_2, \boldsymbol{\mu}, \mathbf{r}\}$. Then, we write Eq \ref{eq:EMD-barycenter3} as 
\begin{equation*}
\mathcal{L}(\mathbf{u}, \boldsymbol{\lambda}) = \mathcal{G}(\mathbf{u}) + \langle \boldsymbol{\lambda}, \tilde{\mathbf{K}}\mathbf{b} \rangle
\end{equation*}
where $\boldsymbol{\lambda} = [\boldsymbol{\lambda}_1;\boldsymbol{\lambda}_2]$, $\mathbf{K} = [\mathbf{D}, \mathbf{I}, \mathbf{-I}, \mathbf{-I}]$, $\tilde{\mathbf{K}} = [\mathbf{K}, \mathbf{0}; \mathbf{0}, \mathbf{K}]$, $\mathbf{b} = [\mathbf{b}_1;\mathbf{b}_2], \mathbf{b}_1= [\text{vec}(\mathbf{M}_1);\boldsymbol{\mu}'_1;\boldsymbol{\mu};\boldsymbol{r}_1]; \mathbf{b}_2 = [\text{vec}(\mathbf{M}_2);\boldsymbol{\mu}'_2;\boldsymbol{\mu};\boldsymbol{r}_2]$. The function $\mathcal{G}= \lVert \mathbf{M}_1 \rVert_{2,1} + \lVert \mathbf{M}_2 \rVert_{2,1} + \alpha(\lVert \boldsymbol{r}_1 \rVert_1 + \lVert \boldsymbol{r}_2 \rVert_1) + \frac{1}{2\rho}(\lVert \boldsymbol{\mu}_1' - \boldsymbol{\mu}_1 \rVert_2 + \lVert \boldsymbol{\mu}_2' - \boldsymbol{\mu}_2 \rVert_2)+\beta\lVert \boldsymbol{\mu} \rVert_1$ is convex and $\tilde{\mathbf{K}}$ is a linear operator. These conditions satisfy Theorem 1 of \cite{chambolle2011first}. If $\lambda_{max}(\boldsymbol{\nabla}^2)$ is the max eigenvalue of $\mathbf{D}\mathbf{D}^{\top}$, then the max eigenvalue of $[\mathbf{D}, \pm \mathbf{I}][\mathbf{D}, \pm \mathbf{I}]^{\top}$ is $\lambda_{\max}(\boldsymbol{\nabla}^2) + 1$. Similarly, for $\mathbf{K}\mathbf{K}^{\top}$, it is $\lambda_{\max}(\boldsymbol{\nabla}^2) + 3$. Since $\tilde{\mathbf{K}}$ is obtained from $\mathbf{K}$ by padding zeros only, $\tilde{\mathbf{K}}$ has the same max eigenvalue as that of $\mathbf{K}$. 
Further, since $\lVert \tilde{\mathbf{K}}\tilde{\mathbf{K}}^{\top} \rVert^2_2 \geq \lambda_{\max}(\tilde{\mathbf{K}}\tilde{\mathbf{K}}^{\top}) = \lambda_{\max}(\boldsymbol{\nabla}^2) + 3$, we can also write the convergence criteria as $\tau_1 \tau_2 \lVert \tilde{\mathbf{K}}\tilde{\mathbf{K}}^{\top} \rVert^2_2  < 1$.

Since we solve for the Lagrangian dual function, we analyse 
the primal dual gap which is given as \cite{jacobs2019solving}
\begin{equation}
    \mathbb{G}(\mathbf{u}, \boldsymbol{\lambda}) = {\sup_{\lVert \boldsymbol{\lambda}' - \boldsymbol{\lambda}_0 \rVert \leq R_1}} \mathcal{L}(\mathbf{u},\boldsymbol{\lambda}') - {\inf_{\lVert \mathbf{u}' - \mathbf{u}_0 \rVert \leq R_2}} \mathcal{L}(\mathbf{u}',\boldsymbol{\lambda}) \nonumber   
\end{equation}
\noindent\textbf{Theorem 2.} Suppose the step sizes $\tau_1$ and $\tau_2$ satisfy $\tau_1 \tau_2\lVert \tilde{\mathbf{K}}\tilde{\mathbf{K}}^{\top} \rVert^2_2 < 1$. Let $\mathbf{u}^N = \frac{1}{N}\sum^N_{n=1}\mathbf{u}_n$ and $\boldsymbol{\lambda}^N = \frac{1}{N}\sum^N_{n=1}\boldsymbol{\lambda}_n$, where $\mathbf{u}_n$ and $\boldsymbol{\lambda}_n$ are sequences generated from Eqns \ref{eq:solveforM} - \ref{eq:solveforlambda}. Then after $N$ iterations, we have,
\begin{equation}
\mathbb{G}(\mathbf{u}, \mathbf{\boldsymbol{\lambda}}) \leq {\sup_{\mathbf{u}, \mathbf{\boldsymbol{\lambda}}}} \frac{1}{2N}  \Bigg(\frac{\lVert \mathbf{u} - \mathbf{u}_0 \rVert_2}{\tau_1} + \frac{\lVert \mathbf{\boldsymbol{\lambda}} - \mathbf{\boldsymbol{\lambda}}_0 \rVert_2}{\tau_2}\Bigg) \nonumber
\end{equation} 
This rate is similar to convergence rates in gradient descent and shows that the gap converges with rate $\mathcal{O}(1/N)$.
 For brevity, we omit the proof and it can be derived as an extension of Theorem 1 \cite{chambolle2011first}.

\subsection{Mutual Information}
In order to understand the underlying reason behind the performance of our method, we provide more insights using mutual information (MI). We first note that the MI between two random variables is given by $\mathcal{I}(X,y) = \mathcal{H}(\mathcal{P}(y)) - \underset{\mathcal{P}(x)}{\mathbb{E}}[\mathcal{H}(\mathcal{P}(y|X))]$. In our case, we take the random variables as model parameters $\boldsymbol{\theta}$ and softmax output $\mathbf{y}$. Then, given a sample $\mathbf{x}$ and dataset $\mathcal{D}$,
\begin{equation}
  \mathcal{I}(\boldsymbol{\theta},\mathbf{y} | \mathcal{D},\mathbf{x}) =   \mathcal{H}(p(\mathbf{y}|\mathbf{x},\mathcal{D})) - \underset{p(\boldsymbol{\theta}|\mathcal{D})}{\mathbb{E}} \mathcal{H}(p(\mathbf{y}|\mathbf{x},\boldsymbol{\theta}))
  \label{eq:mi}
\end{equation}

Eq \ref{eq:mi} measures the information shared between $\boldsymbol{\theta}$ and $\mathbf{y}$. A tractable way of computing $\mathcal{I}(\boldsymbol{\theta},\mathbf{y} | \mathcal{D},\mathbf{x})$ is given in \cite{smith2018understanding}, \cite{houlsby2011bayesian}.
\begin{equation}
    {I}(\boldsymbol{\theta},\mathbf{y} | D,\mathbf{x}) = \frac{1}{C}\sum^C_{j=1}\frac{1}{n}\sum^n_{i=1}(p_{ij} - \hat{\mathbf{p}})^2
\end{equation}
where, $\hat{\mathbf{p}} \in [0,1]^C$ is computed as the mean of all softmax probabilities, $C$ is the number of classes, $\mathbf{p}_{i} \in [0,1]^C$, $p_{ij} \in [0,1]$ denotes the softmax probability for a particular class $j$. A higher ${\mathcal{I}}$ indicates that knowing $\boldsymbol{\theta}$ (or $\mathbf{y}$) gives a higher information about $\mathbf{y}$ (or $\boldsymbol{\theta}$). In other words, the model will perform better if the mutual information is high. 

In addition, we also compute MI between the predictions for the following two cases - (i) clean test set and adversarial test set, and (ii) barycenter of clean test set and barycenter of adversarial test set using \cite{ji2019invariant}. Given a model $f$ parameterised by $\boldsymbol{\theta}$, clean sample $\mathbf{x}_i$ and its adversarial counterpart $\mathbf{x}^{\prime}_i$, the joint probability distribution between natural and adversarial samples is given by the following $C \times C$ matrix,

\begin{equation}
\mathcal{I}(f(\mathbf{x}_i,\boldsymbol{\theta}),f(\mathbf{x}^{\prime},\boldsymbol{\theta})) = \sum^C_{y=1}\sum^C_{y'=1}\mathcal{P}_{yy'} \ln \frac{\mathcal{P}_{yy'}}{\mathcal{P}_{y}\mathcal{P}_{y'}}
\label{eq:mutualinfo}
\end{equation}
where, $\mathcal{P}_{yy'}$ is given as,
\begin{equation}
    \mathcal{P}_{yy'} = \frac{1}{n}\sum^n_{i=1}f(\mathbf{x}_i,\boldsymbol{\theta})f(\mathbf{x}^{\prime}_i,\boldsymbol{\theta})^{\top}
\end{equation}
and the marginals $\mathcal{P}_{y},\mathcal{P}_{y'}$ are obtained by row and column sum of $\mathcal{P}_{yy'}$. A higher value of $\mathcal{I}(.,.)$ indicates that knowing about clean samples gives a higher amount of information about the adversarial samples.

\section{Experiments}
We present elaborate experimental results on CIFAR-10, CIFAR-100 and Tiny ImageNet. 
We use strong baselines of LAS \cite{jia2022adversarial}, LBGAT \cite{cui2021learnable}, PGD-AT \cite{madry2017towards}, TRADES \cite{zhang2019theoretically}, RST \cite{carmon2019unlabeled}. We compare against several popular adversarial training models, MART \cite{wang2019improving}, AWP-A2 \cite{xu2022a2}, RST-RWT \cite{yu2022robust}, TRADES$_{\text{AWP}}$ \cite{wu2020adversarial}, AWP \cite{wu2020adversarial}, LAS$_{\text{AT}}$, LAS$_{\text{TRADES}}$, LAS$_{\text{AWP}}$ \cite{jia2022adversarial}. We also compare with adaptive test time defenses Hedge$_{\text{RST}}$ \cite{wu2021attacking} and TRADES$_{\text{SSL}}$ \cite{mao2021adversarial}. In the Tables, we use \enquote{$+$B} to indicate the results obtained using our approach.

\subsection{Implementation details}
In case of CIFAR10 and CIFAR100, WideResNet34-10 is used and for Tiny ImageNet PreActResnet18 is used. Additionally, we evaluate on CIFAR-10 with WideResNet28-10, WideResNet32-10, WideResNet70-16 and on CIFAR-100 with WideResNet34-20. We use these models for a fair comparison with existing works as these models are widely used for adversarial defense evaluation.
We evaluate against different attacks namely FGSM, PGD-10, PGD-20, CW, and AA using $l_{\infty}$ attack with $\epsilon = 8/255$. Our evaluation protocols are similar to the protocols given in \cite{zhang2019theoretically}, \cite{jia2022adversarial}. We would like to emphasize that we use the checkpoints from the baseline models and perform a single epoch training using clean barycenters. Upon increasing the number of epochs, the clean accuracy improves, however, the adversarial accuracy becomes comparable to that of baseline and further increasing epochs leads to subsequent drop in accuracy against adversarial samples. We use SGD optimizer with a learning rate of 1e-4, momentum = 0.9 without any weight decay. 

In order to compute the barycenter, we set $\rho = 5e-1, \tau_1 = 1e-1, \tau_2 = \alpha = \beta = 1$ and iterations is set to 200. While one can also attack the barycenter, we give experiments for the case where the clean image is attacked. This is because the barycenter itself lies at an $\epsilon$ which is greater than attacker' budget. Thus attacking barycenter has little incentive as in that case the attacked image will lie at an $\epsilon$ outside the given $\epsilon = 8/255$ for the $l_{\infty}$ attack.

\subsection{Comparison on CIFAR-10}
In Table \ref{tab:cifar-10}, we observe that clean performance is better for the models trained with barycenters - TRADES$_{\text{AWP}}+$B, LBGAT$+$B, LAS$_{\text{TRADES}}+$B and RST$+$B. Amongst WRN-28-10 models, RST has the best clean performance and our method enhances it by 1\%. In PGD-10, there is a rise of 2.57\%. In case of AA, there is a boost of 6.49\%. 

In case of WRN-34-10, LAS$_{\text{AT}}+$B shows a huge boost of 10\% under AA. Further, LAS$_{\text{AWP}}+$B shows the best performance under PGD-10, PGD-20 and CW attack amongst WRN-34-10 models. Under AA it shows an improvement of 7.71\%.

\noindent \textbf{Comparison with Adversarial Purification models}
Our RST$+$B model outperforms Hedge$^*_{\text{RST}}$ under all the cases. Against AA, our approach gives 3.1\% higher accuracy compared to Hedge$^*_{\text{RST}}$. We also see that compared to TRADES$_{\text{SSL}}$, TRADES$_{\text{AWP}}$ has a better performance. 

\begin{table}[ht]
\centering
\renewcommand{\tabcolsep}{4.5pt}
\caption{CIFAR-10. $^*$ indicates that the model uses WRN-28-10. \textbf{Bold} font is used to indicate the best performance amongst WRN-34-10 and \textcolor{red}{Red} color font is used to indicate the best performance amongst WRN-28-10.}
\begin{tabular}{l|l|l|l|l|l} 
\hline
Method  & Clean & PGD10  & PGD20 & CW & AA\\ 
\hline
\rowcolor{LightCyan}
\multicolumn{6}{l}{Adversarial Training}  \\
PGD-AT & 85.17 & 56.07 & 55.08& 53.91 &51.69\\
TRADES &85.72 &56.75& 56.10& 53.87 &53.40 \\
MART &84.17& 58.98 &58.56 &54.58& 51.10\\
AWP-A2& 87.54 & -&59.50 &57.42 & 54.86\\
RST-RWT$^*$ & 88.87 & - &64.11 & 62.03 &60.36\\
\hline
\rowcolor{LightCyan}
\multicolumn{6}{l}{Adversarial Purification}  \\
TRADES$_{\text{SSL}}$ & 82.12 & - & - & -& 60.67 \\
Hedge$^*_{\text{RST}}$ & 88.64& - & - & 73.89 & 63.10\\ 
\hline
\rowcolor{LightCyan}
\multicolumn{6}{l}{Adversarial and Barycentric Training}  \\
TRADES$_{\text{AWP}}$ &85.36 & 59.58& 59.25 &57.07 &56.17\\
$+$B & 87.32 & 62.60 &62.32 & 75.85 &65.32\\
\hline
LBGAT & 88.22 & 56.25 & 54.60&  54.29 &52.23\\
\multicolumn{1}{l|}{$+$B} & 88.38 & 59.28 & 58.43  &74.61&61.22\\
\hline
LAS$_{\text{AT}}$ & 86.23 & 57.11 &56.41 & 55.54&53.58\\
\multicolumn{1}{l|}{$+$B} &86.21 &61.08 & 60.64& 74.09& \textbf{63.59} \\
\hline
LAS$_{\text{TRADES}}$ & 85.24 & 57.66 &57.07 &55.45&54.15\\
\multicolumn{1}{l|}{$+$B} &86.15 & 60.32& 60.03  & 73.75&63.43\\
\hline
LAS$_{\text{AWP}}$ &   \textbf{87.74} & 61.09 & 60.16    & 58.22&55.52\\ 
$+$B &   87.45 & \textbf{63.66} & \textbf{61.16}   & \textbf{74.81}& 63.23\\ 
\hline
RST$^*$ &   89.69 & 63.48 & 62.51    & 61.06& 59.71\\ 
$+$B$^*$ & \textbf{\textcolor{red}{90.68}}& 65.12 & 64.38& \textbf{\textcolor{red}{77.08}} & \textbf{\textcolor{red}{66.20}}\\ 
\hline
\end{tabular}
\label{tab:cifar-10}
\end{table}

\subsection{Comparison on CIFAR-100}
In Table \ref{tab:cifar-100}, we observe that our method gives a significant boost under all the cases. In case of strong baseline LAS$_{\text{AWP}}$, our method increases the performance by 0.85\% under clean accuracy. For PGD-20, there is a rise of 0.91\%. In case of CW, there is an increase  of 18.35\%. In other models such as LBGAT, we see a rise of 5.4\% in clean accuracy.

\begin{table}[ht]
\centering
\caption{CIFAR-100 WRN-34-10.}
\begin{tabular}{l|l|l|l|l} 
\hline
Method  & Clean & PGD-10 & PGD-20  & CW \\ 
\hline
\hline
PGD-AT & 60.89 &32.19& 31.69&30.10\\
TRADES & 58.61& 29.20 &28.66&27.05\\
\hline
TRADES$_{\text{AWP}}$ & 60.17 & 33.81 & 33.6& 57.07\\
$+$B & 63.67 & 36.34 & 36.15  &51.92 \\
\hline
LBGAT & 60.64 & 35.13 & 34.53& 30.65\\
$+$B& 66.04 & 36.29 & 36.01 & 52.92\\
\hline
LAS$_{\text{AT}}$& 61.8 & 33.27 & 32.83&31.12 \\
$+$B & 62.45 & 36.60 & 36.17  & 49.60\\
\hline
LAS$_{\text{TRADES}}$ & 60.62 & 32.82 & 32.51  &29.51\\
$+$B &62.58 & 35.22 & 34.96  &50.99\\
\hline
LAS$_{\text{AWP}}$ &   64.89 & 37.11 &36.36 &   33.92 \\ 
$+$B & \textbf{65.50} & \textbf{37.55} & \textbf{37.27} & \textbf{52.27} \\
\hline
\end{tabular}
\label{tab:cifar-100}
\end{table}

\subsection{Comparison on Tiny ImageNet}
We present the results in Table \ref{tab:preactresnet}. In comparison to baselines, our method shows significant improvement in all cases. For LAS$_{\text{AWP}}$, our method improves the performance under clean samples by 1.65\%. In case of PGD-50, our method shows a rise of 1.28\%, and in case of CW attack, our method almost doubles the accuracy. Under AA, LAS$_{\text{TRADES}}$ observes a maximum performance rise by 11.51\%.

\begin{table}[ht]
\centering
\caption{Tiny ImageNet PreActResNet18.}
\begin{tabular}{l|l|l|l|l} 
\hline
Method  & Clean & PGD-50  & CW & AA \\ 
\hline
LAS$_{\text{AT}}$ & 44.86 & 22.16 & 18.54 & 16.74\\
$+$B & 45.12 & 24.54& 37.14 & 27.78\\
\hline
LAS$_{\text{TRADES}}$ & 41.38 & 18.36& 14.50 & 14.08 \\
$+$B& 43.07 & 19.25 & 35.13 & 25.59\\
\hline
LAS$_{\text{AWP}}$ & 45.26 & 23.42 & 19.88 & 18.42 \\
$+$B& \textbf{46.91} & \textbf{24.70} & \textbf{37.93} & \textbf{27.00}\\
\hline
\end{tabular}
\label{tab:preactresnet}
\end{table}

\subsection{Comparison with Curriculum based AT}
In Table \ref{tab:cifar-10-WRN32}, we compare against curriculum based AT methods like CAT \cite{cai2018curriculum}, FAT \cite{zhang2020attacks} and DART \cite{wang2019dynamic}. Under FGSM, PGD-20 and CW, our model shows a huge improvement. In FGSM, our model gives a boost of 4.46\% over FAT. In PGD-20 the boost is 10.89\% and in CW there is a rise of 24.54\%. In case of clean samples, we see that the accuracy of FAT$+$B compared to FAT is less. This may be due to the fact that FAT employs curriculum learning in the training whereas our method does not use curriculum learning. 

\begin{table}[h]
\centering
\caption{CIFAR-10 WRN-32-10.}
\begin{tabular}{l|l|l|l|l} 
\hline
Method  & Clean& FGSM & PGD-20  & CW \\ 
\hline
CAT & 77.43 & 57.17& 46.06 & 42.48 \\
DART & 85.03 & 63.53 & 48.70 & 47.27 \\
\hline
FAT  & \textbf{89.34}&  65.52  & 46.13&  46.82\\
$+$B & 84.59 & \textbf{69.98} & \textbf{57.02} & \textbf{71.36}\\
\hline
\end{tabular}
\label{tab:cifar-10-WRN32}
\end{table}

\subsection{Analysis using Deeper and Wider Models}
We use WRN-70-16 and WRN-34-20 to analyse the effect when the models get deeper and wider. In particular, for clean samples, we can observe that the deeper and wider models give a better boost. In CIFAR-10, WRN-70-16 gives 88.87\% for clean samples which is 3.21\% better than LAS$_{\text{AWP}}$ model' 85.66\%. In contrast, for WRN-34-10, our method gives accuracy similar to that of LAS$_{\text{AWP}}$. 

In CIFAR-100, our method boosts the performance by 8.34\% under AA.
In other cases also we see that the barycenters improve the performance by a significant margin.

\begin{table}[ht]
\centering
\renewcommand{\tabcolsep}{5pt}
\caption{CIFAR-10 (C-10) WRN-70-16 and CIFAR-100 (C-100) WRN-34-20.}
\begin{tabular}{l|l|l|l|l|l} 
\hline
Dataset  & Method  & Clean& FGSM & CW & AA\\ 
\hline
C-10   & LAS$_{\text{AWP}}$ & 85.66 & 70.25 &  58.44 & 57.61\\
 &+B& \textbf{88.87} & \textbf{74.04} & \textbf{75.40} & \textbf{62.54}\\
\hline
C-100 & LBGAT & 62.55 & 43.16 &  31.72 &31.92 \\
& +B& \textbf{66.86} &\textbf{50.92}& \textbf{54.19}& \textbf{40.26}\\ 
\hline
\end{tabular}
\label{tab:cifar-10-WRN70}
\end{table}

\subsection{TSNE} 
\begin{figure}[ht]
     \centering
     \begin{subfigure}[t]{.11\textwidth}
        \includegraphics[width=\textwidth]{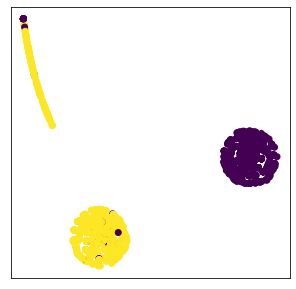}
        \caption{Clean}
    \end{subfigure}
    \begin{subfigure}[t]{.11\textwidth}
        \includegraphics[width=\textwidth]{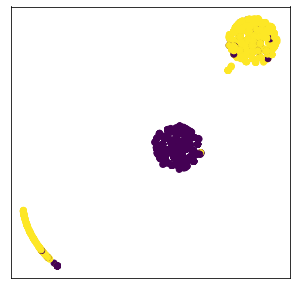}
        \caption{Barycenter}
    \end{subfigure}
    \begin{subfigure}[t]{.11\textwidth}
         \includegraphics[width=\textwidth]{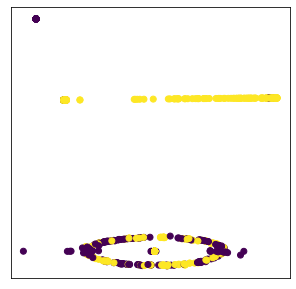}
         \caption{Attacked}
    \end{subfigure}
    \begin{subfigure}[t]{.11\textwidth}
        \includegraphics[width=\textwidth]{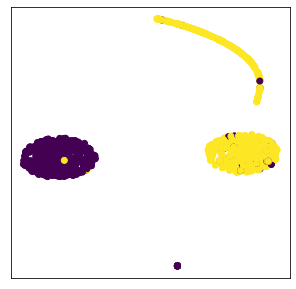}
        \caption{Adv.+Bary.}
    \end{subfigure}    
        \caption{Left to right: Plot of 2D features of Clean image, Barycenter of clean image, Attacked image, Barycenter of adversarial image. MNIST model obtains 51\% accuracy and has only 2D feature vector before the classification layer.} 
        \label{fig:tsne}
    \end{figure}

In Figure \ref{fig:tsne}, we show the tsne plots for MNIST testset with classes 0 and 1. Here, we use a weak MNIST model which has only two dimensions before the classification layer. We deliberately chose a weak model so that we can easily show the effect in low dimensions. Though higher dimensions could be taken, the effect cannot be easily seen due to a highly non-linear transformation from high to low dimension of tsne. We can see that the two clusters yellow and purple are well separated for clean and barycenters of clean images. In case of adversarial samples, the points overlap on each other. However, when we take barycenter of adversarial samples, we again see that the clusters are well separated, similar to the case of clean images. Thus, it is evident that the barycenter nullifies the effect of adversarial noise.  
\subsection{Mutual Information}
In Table \ref{tab:mutualinfo}, we present the study of mutual information. We use LAS$_{\text{AT}}$ and LAS$_{\text{TRADES}}$ on CIFAR-10. The MI is computed using Eq \ref{eq:mi} and Eq \ref{eq:mutualinfo}. Here we note that the MI for LAS$_{\text{AT}}+$B is more for training set compared to that of LAS$_{\text{AT}}$. This indicates that the information available about the labels given the model parameters is high and in turn gives a better clean accuracy. In case of adversarial samples too, we see that the MI is higher for our case. This indicates that the model has better prediction for these samples. Further, the MI for test set is smaller compared to training set which is expected as the model carries more information about train set compared to test set.
\begin{table}[ht]
\centering
\caption{CIFAR-10 WRN-34-10.}
\begin{tabular}{l|l|l|l|l} 
\hline
Method  & Train &Test  & FGSM & CW \\ 
\hline
LAS$_{\text{AT}}$ & 0.029& 0.026&0.020 & 0.019\\
+B & \textbf{0.034}& \textbf{0.029} & \textbf{0.023} & \textbf{0.022}\\
\hline
LAS$_{\text{TRADES}}$&0.048 & 0.040& 0.033& 0.032\\
+B&\textbf{0.054} & \textbf{0.045} & \textbf{0.037} & \textbf{0.036}\\
\hline
\end{tabular}
\label{tab:mutualinfo}
\end{table}

In Table \ref{tab:mutualinfo2}, we present the results obtained using Eq \ref{eq:mutualinfo}. We observe that for the model trained with barycenter, the MI is higher between the barycenter of clean and adversarial samples. Thus, the model does better on barycenter of adversarial samples compared to baseline LAS$_{\text{AT}}$ and TRADES$_{\text{AWP}}$. This is consistent across FGSM, PGD-10 and CW attacks.
\begin{table}[ht]
\centering
\caption{Mutual Information for CIFAR-10 WRN-34-10.}
\begin{tabular}{l|l|l|l} 
\hline
Method  &  FGSM & PGD-10& CW \\ 
\hline
LAS$_{\text{AT}}$ & 0.218 & 0.198&0.203\\
+B & \textbf{0.275} & \textbf{0.241} & \textbf{0.264} \\
\hline
LAS$_{\text{TRADES}}$&0.576& 0.554& 0.563\\
+B& \textbf{0.629}& \textbf{0.572}& \textbf{0.618}\\
\hline
\end{tabular}
\label{tab:mutualinfo2}
\end{table}

\subsection{Sensitivity to Barycenter Parameters}
\begin{figure}[ht]
     \centering
         \includegraphics[width=1.2cm, height=1.2cm, cfbox=blue 1pt 1pt]{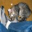}
        \includegraphics[width=1.2cm, height=1.2cm, cfbox=red 1pt 1pt]{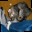}
        \includegraphics[width=.07\textwidth]{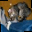}
        \includegraphics[width=.07\textwidth]{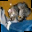}
        \includegraphics[width=.07\textwidth]{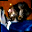}\\
        \includegraphics[width=.09\textwidth]{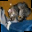}
        \includegraphics[width=1.5cm, height=1.5cm, cfbox=red 1pt 1pt]{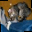}
        \includegraphics[width=.09\textwidth]{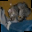}
        \includegraphics[width=.09\textwidth]{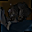}\\
         \includegraphics[width=1.5cm, height=1.5cm, cfbox=red 1pt 1pt]{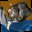}
          \includegraphics[width=.09\textwidth]{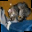}
           \includegraphics[width=.09\textwidth]{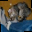}
            \includegraphics[width=.09\textwidth]{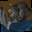}\\
        \caption{Top row: Blue boundary represent the given image. Barycenter for iterations = 200, $\tau_1=1e-1$, $\tau_2$ = {1, 1e-1, 1e-2, 1e-3}. Second: barycenter for iterations = 200, $\tau_1$={1, 1e-1, 1e-2, 1e-3}, $\tau_2 =1$. Third: iterations = {200, 100, 50, 10}, $\tau_1$=1e-1, $\tau_2$=1. The red boundary indicates the images obtained from default settings of the parameter which are used for all experiments.}
        \label{fig:barycenter-sensitivty}
\end{figure}

In Figure \ref{fig:barycenter-sensitivty}, we demonstrate the sensitivity to different parameters involved in the computation of barycenter. In the top row, we fix the number of iterations to 200 and $\tau_1 = 1e-1$. Here we observe that increasing $\tau_2$ makes the barycenter brighter. In the second row, increasing $\tau_1$ makes the barycenter darker. Decreasing iterations has a similar effect in the last row. We see that unless there is a change of order of magnitude, the appearance does not substantially change. Thus, our proposed Beckman barycenter is robust with respect to the parameter settings.

\begin{figure}[hb]
     \centering
      \begin{subfigure}[h]{.12\textwidth}
        \includegraphics[width=\textwidth]{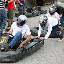}
        \caption{Clean}
    \end{subfigure}
    \begin{subfigure}[h]{.12\textwidth}
        \includegraphics[width=\textwidth]{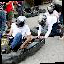}
        \caption{Barycenter}
    \end{subfigure}
    
    \begin{subfigure}[h]{.12\textwidth}
        \includegraphics[width=\textwidth]{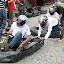}
        \caption{Adversarial}
    \end{subfigure}
    \begin{subfigure}[h]{.12\textwidth}
        \includegraphics[width=\textwidth]{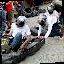}
        \caption{Adv.+Bary.}
    \end{subfigure}
        \caption{Illustration of barycenters from Tiny ImageNet.} 
        \label{fig:barycenter-samples}
    \end{figure}

The Figure \ref{fig:barycenter-samples} demonstrates a visualization of clean image, its barycenter, attacked image and its barycenter. The difference between them is imperceptible except for the black portion in the border for barycenters. This arises due to interpolation of the images from rotation.

\section{Conclusion}
In this work we introduce Beckman barycenter analogous to Wasserstein barycenter. We use the Beckman OT formulation and analytically solve for the barycenter. Defining the baycenter using Beckman OT also has the advantage that the computational tools to obtain barycenter are well known and efficient. This overcomes the complexity in solving Wasserstein barycenters. Further, we show that barycenter can be used for enhancing the performance of adversarially trained models. Our training is very efficient as we only need a single epoch. Experimental analysis demonstrates state-of-art results against different attacks.

\clearpage
\bibliographystyle{named}
\bibliography{ijcai22}

\end{document}